\documentclass[fleqn,10pt]{wlscirep}
\usepackage[utf8]{inputenc}
\usepackage[T1]{fontenc}
\usepackage{amsmath}

\DeclareMathOperator*{\argmin}{arg\,min}
\newcommand\norm[1]{\left\lVert#1\right\rVert}

\title{Using interpretable boosting algorithms for modeling environmental and agricultural data}

\author[1, 2]{Fabian Obster}
\author[2]{Christian Heumann}
\author[3]{Heidi Bohle}
\author[3,*]{Paul Pechan}
\affil[1]{University of the Bundeswehr Munich, Department of business administration, Neubiberg, 85577, Germany}
\affil[2]{Ludwig Maximilian University Munich, Department of Statistics, city, 80539, Germany}
\affil[3]{Ludwig Maximilian University Munich, Institute of Communication and Media Research, 80539, Germany}

\affil[*]{paul.pechan@ifkw.lmu.de}

\begin{abstract}
We describe how interpretable boosting algorithms based on ridge-regularized generalized linear models can be used to analyze high-dimensional environmental data. We illustrate this by using environmental, social, human and biophysical data to predict the financial vulnerability of farmers in Chile and Tunisia against climate hazards. We show how group structures can be considered and how interactions can be found in high-dimensional datasets using a novel 2-step boosting approach. The advantages and efficacy of the proposed method are shown and discussed. 
Results indicate that the presence of interaction effects only improves predictive power when included in two-step boosting. The most important variable in predicting all types of vulnerabilities are natural assets. Other important variables are the type of irrigation, economic assets and the presence of crop damage of near farms.
\end{abstract}
\begin{document}

\flushbottom
\maketitle
\thispagestyle{empty}

\section*{Introduction}
 In this work, we show how interpretable boosting algorithms can be used to predict financial vulnerabilities against multiple hazards based on environmental factors but also based on human, social, and biophysical factors as well as their interactions. For finding interactions we propose a new method based on two-step boosting, which is still interpretable and blends together with component-wise boosting. Interpretability tools like variable importance, effect sizes, and partial effects are utilized to better understand the underlying factors that may cause these vulnerabilities against climatic changes. \\
 Model-based boosting algorithms have been used in environmental sciences for multiple purposes. For example for quantifying several soil parameters based on soil samples \cite{li_data_2021}, predicting the financial wellbeing of farmers based on environmental factors \cite{obster_factors_2023}, and predicting the number of zoo visitors based on climatic variables \cite{obster_improving_2023}. Also non-interpretable boosting algorithms based on classification or regression trees like Adaboost \cite{freund_decision-theoretic_1997} have been used for environmental predictions based on environmental data because of their high predictive power. Applications include landslide susceptibility \cite{jennifer_feature_2022} and predicting the presence of juvenile sea-trouts based on environmental factors \cite{froeschke_spatio-temporal_2011}. \\
Through the proposed boosting models we want to achieve the following goals:
\begin{itemize}\label{list:requirenments}
\item \textbf{Predictive Power}: The model should not only have a good fit for the analyzed data but also for unseen data from the same domain assuming a similar distribution of the variables.
\item \textbf{Interpretability}: We are interested in the question of which variables are associated with the outcome. But we also want to know how the associations look like. In the agronomic case, we want to derive actions to reduce vulnerability against adverse environmental changes. This is only possible if the effect of adaptive measures is known. Only if the associations are known, one can state causal hypotheses and test them with new specific experiments. We also want the effects to be modeled as simply as possible while retaining the power of the model. Linear effects should be prioritized over nonlinear effects and over interaction effects. Black-Boxes should be avoided in this case. 
\item \textbf{Sparsity}: We consider high dimensional data sets where the number of variables $p$ is relatively large compared to the number of observations $n$ or even possibly higher if we consider the case with interactions. Out of the many possible variables, we want to know which ones are actually associated with the outcome and which ones are not. Therefore, the model should perform variable selection to enforce sparsity. The goal is to find the smallest subset of variables that still has high predictive power. Sparsity also increases interpretability because the scientist and stakeholders only have to look at the truly relevant variables and can disregard the unimportant ones. In the vulnerability setting this could mean that farmers focus on selected variables like the type of irrigation systems rather than not selected variables like financial adaptive measures.
\item \textbf{Complexity}: The model should be as complex as necessary and as simple as possible. Complexity is the characteristic that balances all previously stated points. Out of two explanations with the same predictive power the model should pick the one that is simpler. By simpler, we mean sparser, more interpretable, and without interactions. On the other hand, we do not want to neglect important complexities like non-linearity and interactions. It is important to identify if some variables are modified by others. There could also be non-hierarchical interactions, where a variable has by itself no effect on the outcome, but may have a positive effect in one subset of the data and a negative one in the other. One example could be, that in one region a high variety of crops has a positive effect on vulnerability and in another region a negative effect.
\item \textbf{Group structure} The variables in the data can be clustered into groups. "Climate change experience" is one example and contains the binary variables "increasing temperature", "increasing drought", "increasing extreme weather" and "decreasing rain". The question is whether the outcome is influenced by each or only by some of the individual variables or if they act as a group due to the similarity. Group structures also increase interpretability, because it is often easier for humans to comprehend the overall effect of an abstract concept than to look at all its facets. 
\end{itemize}
There are many approaches to deal with each of the above specifications. For example, sparsity can be achieved through Lasso Regression \cite{tibshirani_regression_1996} or boosted Lasso \cite{zhao_boosted_2004}, predictiveness can be achieved through a big variety of models and group structures can be incorporated with the sparse group lasso \cite{simon_sparse-group_2013}.\\ In this work we focus on how these goals can be met using boosting algorithms, namely componentwise boosting (mb), componentwise boosting with interactions (mb int), sparse group boosting (sgb), and two-step boosting for interactions (2-boost). We compare their predictive power, effect sizes, and the relative importance of variables/groups. In the following, we describe the used methods for the analysis and discuss how they help to achieve the stated goals using modifications of the generic boosting algorithm. 
\section*{Methods}
\subsection*{Introduction of the data}
The data was collected using focus meetings and face-to-face survey interviews with 400 cherry farmers in Chile and 401 peach farmers in Tunisia. The selection criteria were that farmers manage the farm and derive over 70 percent of their income from farming activities. For the data collection, guidance was sought from our institute (ICMR, LMU) about the survey implementation and data use that included the participation of human subjects. The farm data was collected according to data collection procedures applicable in each country. Informed consent for the data collection was provided by the survey participants. No personality identifiable data was collected, assuring full anonymity. A descriptive description of the data \cite{pechan_climate_2023} and further mixed methods analysis on vulnerability \cite{pechan_reducing_2023} with similar data was performed. The R code of the analysis can be found at https://github.com/FabianObster/boostingEcology.
\subsubsection*{Independent variables}
The analyzed variables can be clustered into groups, including 
\begin{itemize}
    \item Climate experience group (Increasing temperature, decreasing rain, increasing drought, increasing extreme weather)
    \item Natural asset group (geographical regions)
    \item Social asset group (reliance on/use of information, trust in information sources, community, science or religion)
    \item Human asset group (age, gender, education)
    \item Biophysical asset group (farm size, water management systems used on the farm, diversity of crops used, adaptive measures) 
    \item Economic asset group (farm debt, farm performance, reliance on orchard income)
    \item Goals group (Keep tradition alive, work independently)
    \item Harm group (Climate threatens farm, Optimism)
    \item Spatial group (Crop damage near farms, Crop damage of farms in Country)
\end{itemize}
An overview of all variables and the belonging groups can be found in Tables \ref{tab:variables} and \ref{tab:variables_2}. There, also the number of farmers in each category can be found. 
\subsubsection*{Outcome variables}
The outcome variables measure financial vulnerability against the 5 climate hazards, increasing winter temperatures, increasing summer temperatures, decreasing rainfall, increasing drought, and increasing extreme weather based on self-assessment of the farmers.
For each of the hazards, a binary variable indicating if a farmer is vulnerable to the hazard is defined as the outcome variable. The main category includes farmers, who are not financially vulnerable and the reference category includes farmers who are financially vulnerable. The number of farmers in each category can be found in Table \ref{tab:outcomes}. 
\subsection*{Interaction variables}
22 variables were used as variables that may have an interaction effect with the other variables on the outcome. The interaction variables include regions as well as socio-demografic variables amongst others and are indicated in bold in tables \ref{tab:variables} and \ref{tab:variables_2}. Together with all other variables, this yields 1366 interaction terms and over 4000 possible model parameters to estimate. Since there are 801 farmers in the data, finding interactions results in a "p>n" problem, where the number of variables in the design matrix is greater than the number of observations.
\subsection*{General setup, model formulation and evaluation}
All analyses were performed with R \cite{rstudio_team_rstudio_2020} and the boosting models were fitted with the package "mboost" \cite{hofner_model-based_2014}. \\
 Since all outcome variables are binary, we use the Ridge penalized negative log-likelihood of the binomial distribution as a loss function and a logit link, which yields
\begin{align*}
h(\beta, X_i) = P(y_i == 1) = \frac{1}{1+\exp{(-X_i^{T}\beta)}}, \\
    l(y, h) =  -\left[\sum_{i=1}^n y_i \log{(h(\beta, X_i))}+(1-y_i)\log{(1-h(\beta, X_i))}\right] + \lambda \norm{\beta}_2^2.
\end{align*}
Before performing any analysis the data was split into 70 percent training data and 30 percent test data, which was only used for the final evaluation. Variable importance and partial effects were computed using the whole data after the predictive analysis. Model evaluation was based on the area under the receiver operator curve (ROC) and computed using the test data. The area under the ROC (AUC) takes both the true positive and the false positive rate into account by considering all possible thresholds of predicted probabilities computed by a prediction model. \\
In the analysis, we use multiple boosting models for multiple purposes. All boosting models were fitted with the R package "mboost" \cite{hothorn_mboost_2022}. For early stopping, the stopping parameter was determined using a 10-fold cross-validation performed at every boosting step. The first and most simple one is component-wise model-based boosting (mb) with ridge-regularized linear effects of all variables, such that the degrees of freedom are all equal to one. This model allows us to perform variable selection and allows for a comparison between all variables regarding their relative importance. For the second model, we used sparse group boosting with a mixing parameter $\alpha = 0.5$, which balances group selection and individual variable selection. This way it is possible to see if variables are important on their own for the outcome, or if they rather act as groups of variables.\\ To find interactions in the data we use two approaches. The first one is the standard approach by defining linear effects and interaction effects at the same time in each iteration. Then the model can decide weather it selects the main effects or the interaction effects. In the second approach we use a two-stage boosting model. As the first step we use the already fitted mb model, which only uses individual linear base-learners. The second step uses solely interactions. This way linear base-learners are prioritized over interaction base-learners since they are fitted first.
\\ 
This remaining part of the methods section is more technical and may be skipped by the application-oriented reader. 
\subsection*{Generic boosting algorithm}
We will start with the general formulation of the boosting algorithm which can also be described as a functional gradient descent algorithm. The goal is to find a function $f^*$ that minimizes some Loss function $l(y,f)$. Here, we only consider differentiable convex loss functions. The loss function has two arguments. The first argument $y \in \{1,...,n\}$ is the outcome variable with $n$ observations. The second argument $f$ is a real-valued function $f: \mathbb{R}^{n \times p} \mapsto \mathbb{R}$, which is a function of the data $X \in \mathbb{R}^{n \times p}$. 

Another way of fitting sparse regression models is through the method of boosting. The fitting strategy is to consecutively improve a given model by adding a base-learner to it. Throughout this article, we refer to a base-learner as a subset of columns of the design matrix associated with a real-valued function.
To enforce sparsity, each base-learner only considers a subset of the variables at each step \cite{buhlmann_boosting_2007}. In the case of component-wise $\mathcal{L}^2$ boosting, each variable will be a base-learner. 
In the case of a one-dimensional B-Spline, a base-learner is the design matrix representing the basis functions of the B-Spline.
The goal of boosting in general is to find a real valued function that minimizes a typically differentiable and convex loss function $l(\cdot,\cdot)$. Here we will consider the negative log-likelihood as a loss function to estimate $f^*$ as
\begin{equation*}
f^*(\cdot)=\argmin_{f(\cdot)} \mathbb{E}[l(y,f)].
\end{equation*}

\subsection*{General functional gradient descent Algorithm\cite{friedman_greedy_2001}}

\begin{enumerate}\label{algo:boosting}
\item Define base-learners of the structure $h: \mathbb{R}^{n \times p} \to \mathbb{R}$
\item Initialize $m=0$ and $\widehat{f}^{(0)} \equiv 0$ or $\widehat{f}^{(0)} \equiv \overline{y}$
\item Set $m = m+1$ and compute the negative gradient $\frac{\partial}{\partial f} l(y,f)$ and evaluate it at $\widehat{f}^{[m-1]}$. Doing this yields the pseudo-residuals $u_1,...,u_n$ with
\begin{equation*} 
u_i^{[m]} = \frac{\partial}{\partial f} l(y_i,f)|_{f = \widehat{f}^{[m-1]}},
\end{equation*}
for all $i = 1,..., n$
\item Fit the base-learner  $h$ with the response $(u_1^{[m]},...,u_n^{[m]})$ to the data. This yields $\widehat{h}^{[m]}$, which is an approximation of the negative gradient
\item Update
\begin{equation*}
\widehat{f}^{[m]} = \widehat{f}^{[m-1]} + \eta \cdot \widehat{h}^{[m]}
\end{equation*}
here $\eta$ can be seen as learning rate with $\eta \in ]0,1[$
\item Repeat steps 3,4 and 5 until $m=M$
\end{enumerate}
\subsection*{Boosted ridge regression}
The loss function $l(\cdot,\cdot)$ can be set to any function. In the case of interpretable boosting, the negative log-likelihood is a reasonable choice. 
The log-likelihood can also be modified using a Ridge penalty. By introducing the hyperparameter $\lambda>0$, one can modify the loss function $l$. Let $h$ be a function of a parameter vector $\beta \in \mathbb{R}^p$ and the design matrix $X \in \mathbb{R}^{n\times p}$, then
\begin{equation*}
    l_{\text{Rige}}(u,h) = l(u,h)+\lambda\norm{\beta}_2^2
\end{equation*}
is the Ridge penalized loss function.
By increasing $\lambda$, the parameter vector $\beta$ can be shrunken towards zero. Closely related to $\lambda$ are the degrees of freedom. Let  $S$ be the approximated generalized ridge hat matrix as in Proposition 3 in \cite{tutz_boosting_2007}. 
We remark that in the special case of ordinary least squares ridge regression we have $S = X(X^TX+\lambda I)^{-1}X^T$. Generally, the degrees of freedom can be defined as
\begin{equation*}
    \text{df}({\lambda}))= \mathrm{tr}(2S-(S)^TS)).
\end{equation*}
It is recommended to set the regularization parameter for each base-learner, such that the degrees of freedom are equal for all base-learners. Thus, the regularization parameter enables using complex base-learners like polynomial effects and simple effects like linear effects at the same time. Since the more complex base-learners are regularized more than the simpler ones it is possible to prioritize simple and more interpretable base-learners over complex ones, introducing an inductive bias towards interpretability, as we demanded in the problem statement.
\subsection*{Component-wise and group component-wise boosting} 
In step 4 of the general functional gradient descent algorithm, the function $h$ is applied. Instead of just one function, one can also use a set of $R$ functions $\{(h_r)_{r \leq R}\}$. Then the update in step 5 is only performed with the function that has the lowest loss function applied to the data, meaning $r^* =\argmin_{r \leq R} \mathbb{E}[l(u,h_r)]$.
In the case of component-wise boosting, for each variable in the dataset, a function is used that is only a function of this variable and not the others. This way in each step only one variable is selected. Then through early-stopping, or setting $M$ relatively smaller compared to the number of variables in the dataset, a sparse overall model can be fitted. This addresses the sparsity requirement in the problem statement section. In the case of grouped variables, one can also define base-learners as groups of variables, which are a function of only the variables belonging to one group. These could be all item variables that belong to a specific construct like in sociological data \cite{agarwal_verifying_2011} or all climate change-related variables in agricultural and environmental data \cite{obster_factors_2023}. This allows group variable selection, where only a subset of groups is selected, yielding a group/construct-centric analysis rather than on an individual-variable basis. This way, the group structure can be taken into account.

\subsection*{Sparse group boosting}
It is also possible to use individual and group-based base-learners at the same time. Then at each step, either an individual variable or a group of variables is selected. Using a similar idea as in the sparse group lasso \cite{simon_sparse-group_2013}, the sparse group boosting can be defined \cite{obster_sparse-group_2022}. We do this again by modifying the degrees of freedom. Each variable will get its own base-learner, and each group of variables will get one base-learner, containing all variables of the group. Let $G$ be the number of groups and $p_g$ the number of variables in group $g$. Then, for the degrees of freedom of an individual base-learner
$x_j^{(g)} \in \mathbb{R}^{n \times 1}$  we will use
\begin{equation*}
     \text{df}({\lambda}^{(g)}_{j}) = \frac{1}{p_g }\cdot \alpha.
\end{equation*}
For the group base-learner we use
\begin{equation*}
     \text{df}(\Tilde{\lambda}^{(g)}) = \frac{1}{p_g}\cdot (1-\alpha).
\end{equation*}

The mixing parameter $\alpha \in [0,1]$ allows to change the prioritization of groups vs. individual variables in the selection process. If $ \text{df}(\lambda) = 0$ means $\lambda \to \infty$, $\alpha = 1$ yields component-wise boosting, and $\alpha = 0$ yields group boosting.

\subsection*{Two-step boosting}
In the generic boosting algorithm, a single set of functions is applied sequentially to the data. While there is variable selection within the set of functions, the set itself does not change during the boosting procedure. We describe a modification of the general that allows more flexibility, namely a two-step version of boosting. A similar idea of two-step boosting, called hierarchical boosting has been used in genetic research \cite{pybus_hierarchical_2015} in transfer learning \cite{wang_hierarchical_2016}, and also deep learning applications \cite{yang_feature_2019}. In most cases, hierarchical boosting is used, if the outcome variable consists of a hierarchical class structure \cite{valentini_hierarchical_2014}. In contrast to the data analyzed in the literature, the data we analyze here does not contain hierarchical class structures. Hence, we do not use hierarchical boosting as in most cases presented in the literature, but for hierarchical and non-hierarchical interaction detection. \\ We formulate and generalize the two-step boosting.
Let $K$ be the number of steps and for every step $k \leq K$ let $ H_k$ be the set of base-learners. 
\subsubsection*{K-step boosting algorithm}
\begin{enumerate}
    \item Set $K$ as the number of steps
    \item For every step $k \leq K$ define the set of base-learners $ H_k$ to be used and set $M_k$ to the number of boosting iterations
    \item Initialize $m_0=0$ and $\widehat{f}^{(0)} \equiv 0$ or $\widehat{f}^{(0)}$
    \item For $k \leq K$ repeat:
    \item For $m_k \leq M_k$ perform steps 2-6 of the general boosting algorithm 
    \item Set Initialization $m_k=0$ and $u^{[0]} = u^{[M_{k-1}]}$
\end{enumerate}
One may ask why it is necessary to run multiple boosting algorithms after each other if it is possible to just use more base-learners in parallel in the original method. Previous research has shown high predictive powers in some combinations of steps. However, as described in the problem statement for us predictive power is only one part of the requirements and not necessarily desirable if it comes at the cost of interpretability and understanding of the data. Also, the sequential nature of the algorithm reduces computational improvements through parallelization, as not all base-learners can be fitted in the same boosting iteration in parallel. The k-step boosting algorithm can also be seen as a special case of the general boosting algorithm, where the base-learners themselves are boosting algorithms.
\subsection*{Variable importance}
For each of the  previously described boosting methods, it is possible to compute a variable importance measure. In each step, the log-likelihood is computed, which means that one can compute the reduction of log-likelihood attributed to the base-learner being selected in the step. After the fitting for each base-learner the total reduction of likelihood can be computed. This way, one can compute the percentage of reduction in the negative log-likelihood attributed to each base-learner, regardless of the type of base-learner. The variable importance allows us to compare the relative importance of variables compared to each other and is distinct from the concept of significance or p-values which tests a hypothesis of a parameter not being zero based on a set of assumptions. Hence a variable can be important in boosting while not being significant base on classical regression and vice versa. 
\subsection*{Partial effect and effect sizes}
For boosted generalized linear models, partial effects can be computed \cite{hofner_model-based_2014}.
Similar to classical logistic regression, odds ratios for all base-learners can be computed by first summing up all linear predictors for one base-learner. These odds ratios can then be interpreted similarly to effect sizes in logistic regression. Based on the linear predictor one can also compute predicted probabilities for categories of variables if all other base-learners are set to average values. This way partial effects can be plotted, both for individual variable base-learners and for interaction-base-learners. Thus model-based boosting models are by themselves interpretable compared to other machine learning models where only post-hoc explanations can be derived. One can also track which variable was selected in each boosting iteration and thus understand how the model works internally.

\section*{Applications}
\subsection*{Predictability}

\begin{figure}[ht]
\centering
\includegraphics[width=\linewidth]{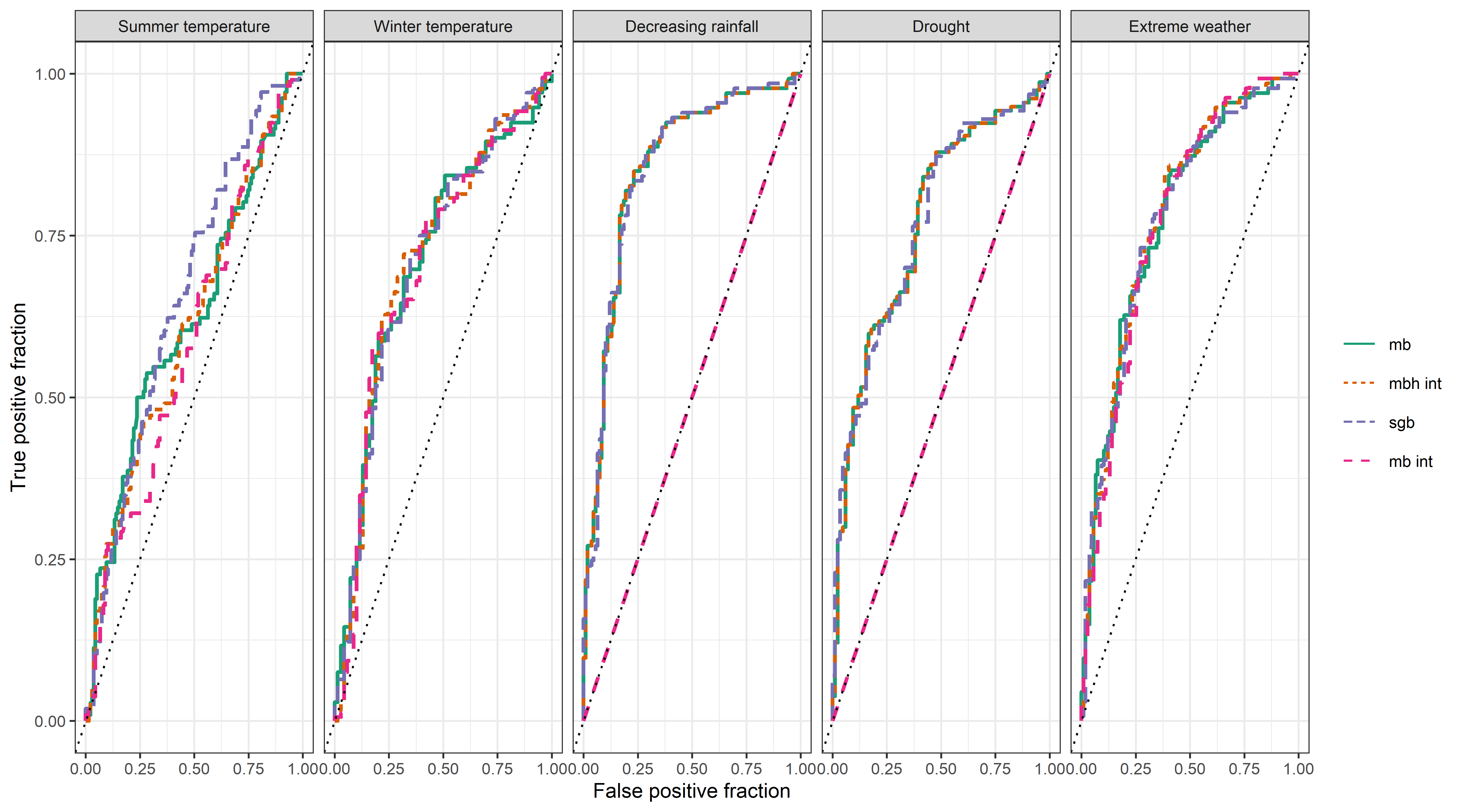}
\caption{ROC-curves for the sparse group boosting (sgb), component-wise boosting (mb), parallel boosting with interaction (mb int) and two-step boosting with interactions (2-boost) for all vulnerability outcomes evaluated on the test data.
\label{fig:auc_all}
}
\end{figure}

Referring to Table \ref{tab:auc_no_vul} and Figure \ref{fig:auc_all} we can see that the AUC values are comparable between the boosting models except for the interaction model with parallel estimation. Averaging the AUC values across the five vulnerabilities, sgb yields 0.752, mb and 2-boost yield 0.745, and mb int 0.613. For precipitation and drought vulnerability, the parallel estimation of interactions resulted in no variables being selected and therefore the AUC takes a value of 0.5. In 2-boost, also no variables were selected in the second estimation resulting in the same model as mb, which had the highest AUC for these outcome vulnerabilities. For summer temperature vulnerability, sgb had the highest AUC, and for winter temperature and extreme weather 2-boost had the highest AUC. Comparing the predictability of the individual vulnerabilities with each other, we see, that vulnerability against decreasing rainfall can be predicted better with the given variables, followed by vulnerability against increasing extreme weather, decreasing drought, increasing winter temperature, and summer temperature.

\subsection*{Importance of individual variables and groups}
\begin{figure}[ht]
\centering
\includegraphics[width=\linewidth]{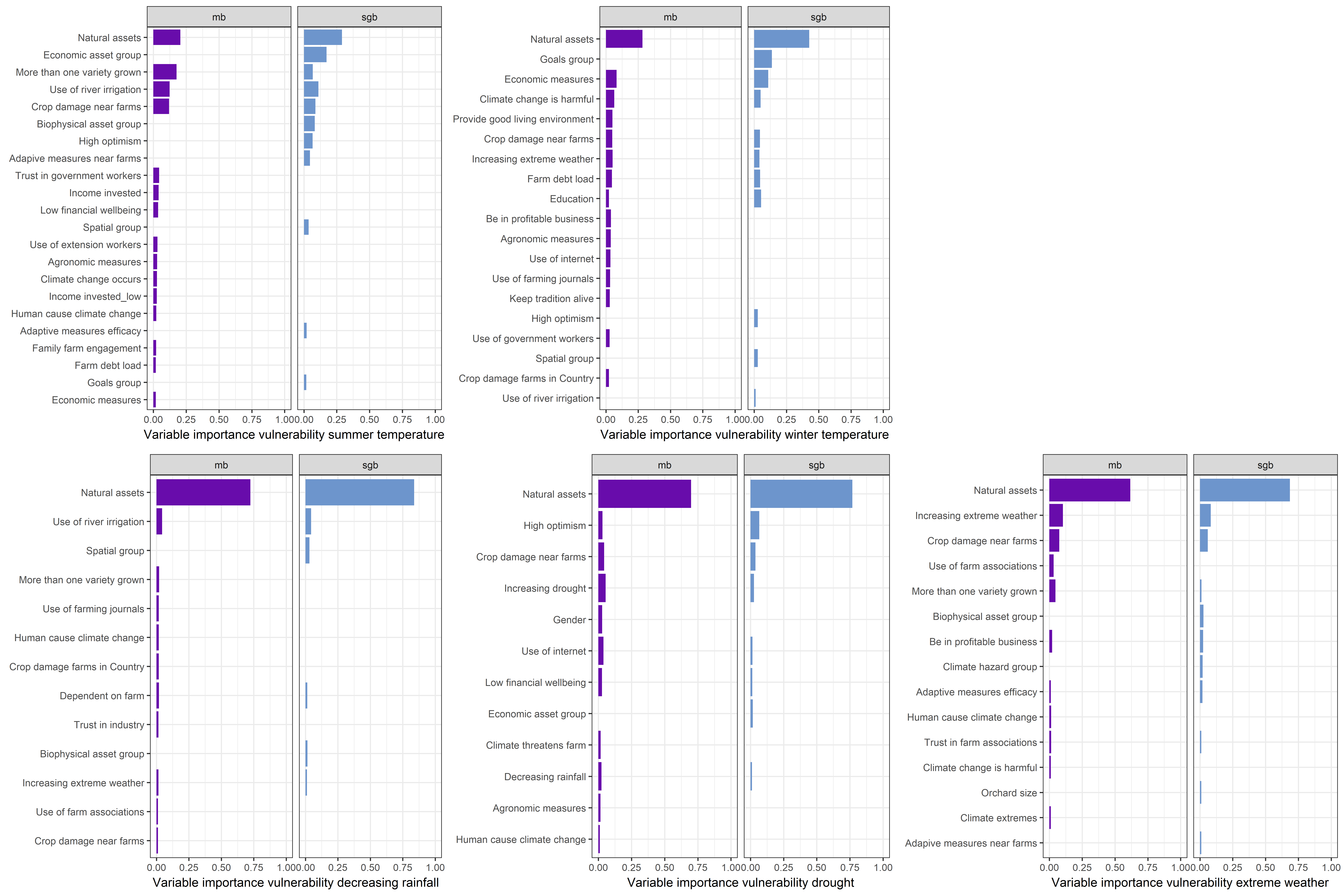}
\caption{Comparison of variable importance based on component-wise boosting (mb) and sparse group boosting (sgb) for each vulnerability. The ordering of variables is based on the sum of relative importance for both models. only variables with a relative contribution of at least one percent and at most 15 variables per model are shown}
\label{fig:varimp}
\end{figure}

Comparing the variable importance of the sparse group boosting (sgb) and componentwise boosting (mb) in Figure \ref{fig:varimp}, it becomes apparent, that while there is some overlap, also some variables differ. The single most important variable for all outcomes is "Natural assets" indicating the four regions of the farm. However, the relative importance of the natural assets is higher for sgb than for mb for all five vulnerabilities. Groups seem to be more important in explaining increasing temperature vulnerability than the other vulnerabilities, as the economic asset group is the second most important variable for summer temperature vulnerability and the goals group is the second most important variable for winter temperature vulnerability. The spatial group is the third most important variable for decreasing rainfall vulnerability but the relative importance is minor compared to the most important variable. 
\subsection*{Importance of interactions}
\begin{figure}[ht]
\centering
\includegraphics[width=\linewidth]{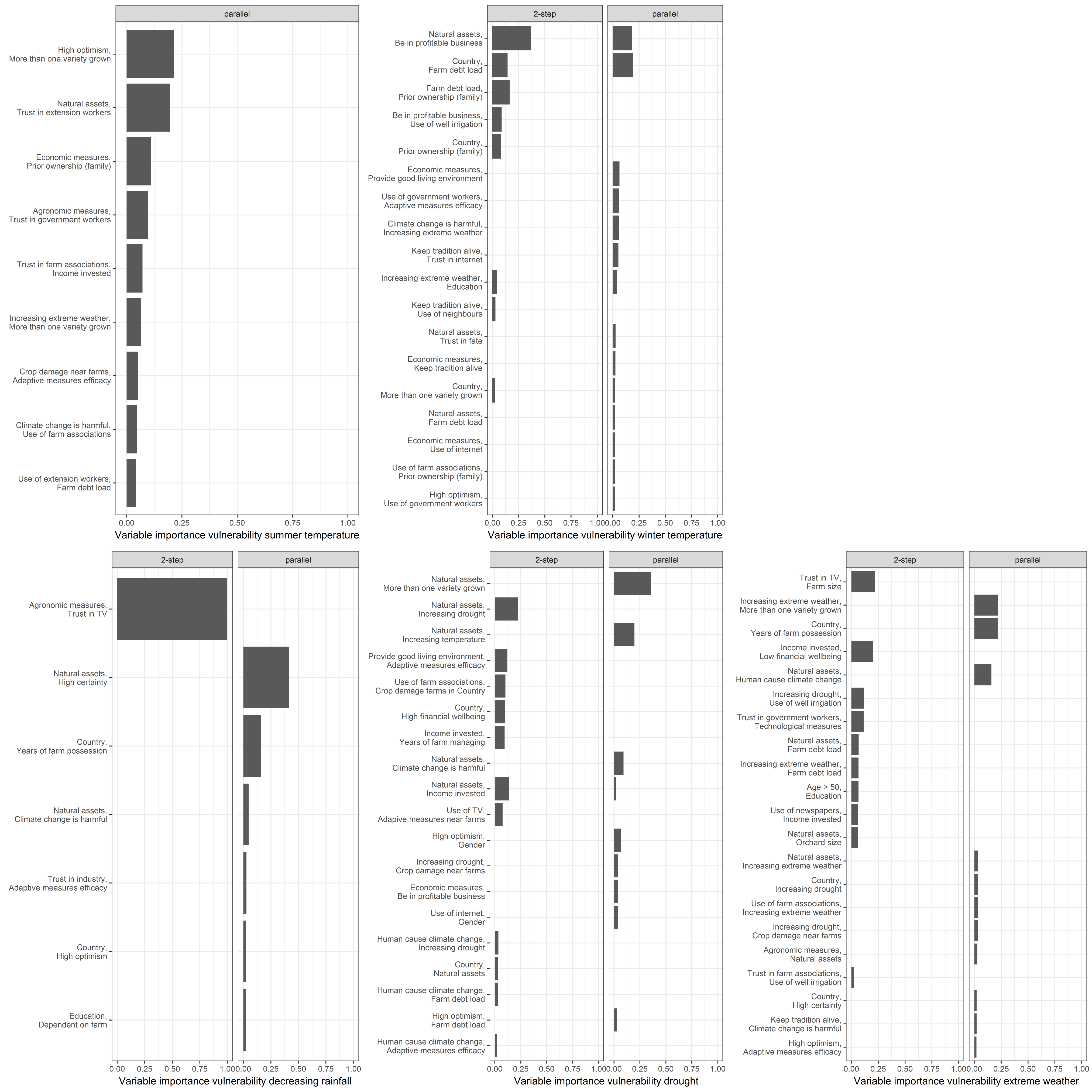}
\caption{Variable importance of interaction terms in two-step estimation (2-boost) and parallel estimation (2-boost) for each vulnerabilities. The ordering of variables is based on the sum of relative importance for both models Only variables with a relative contribution of at least two percent and at most 15 variables per model are shown}
\label{fig:varimp_interact}
\end{figure}

\begin{figure}[ht]
\centering
\includegraphics[width=\linewidth]{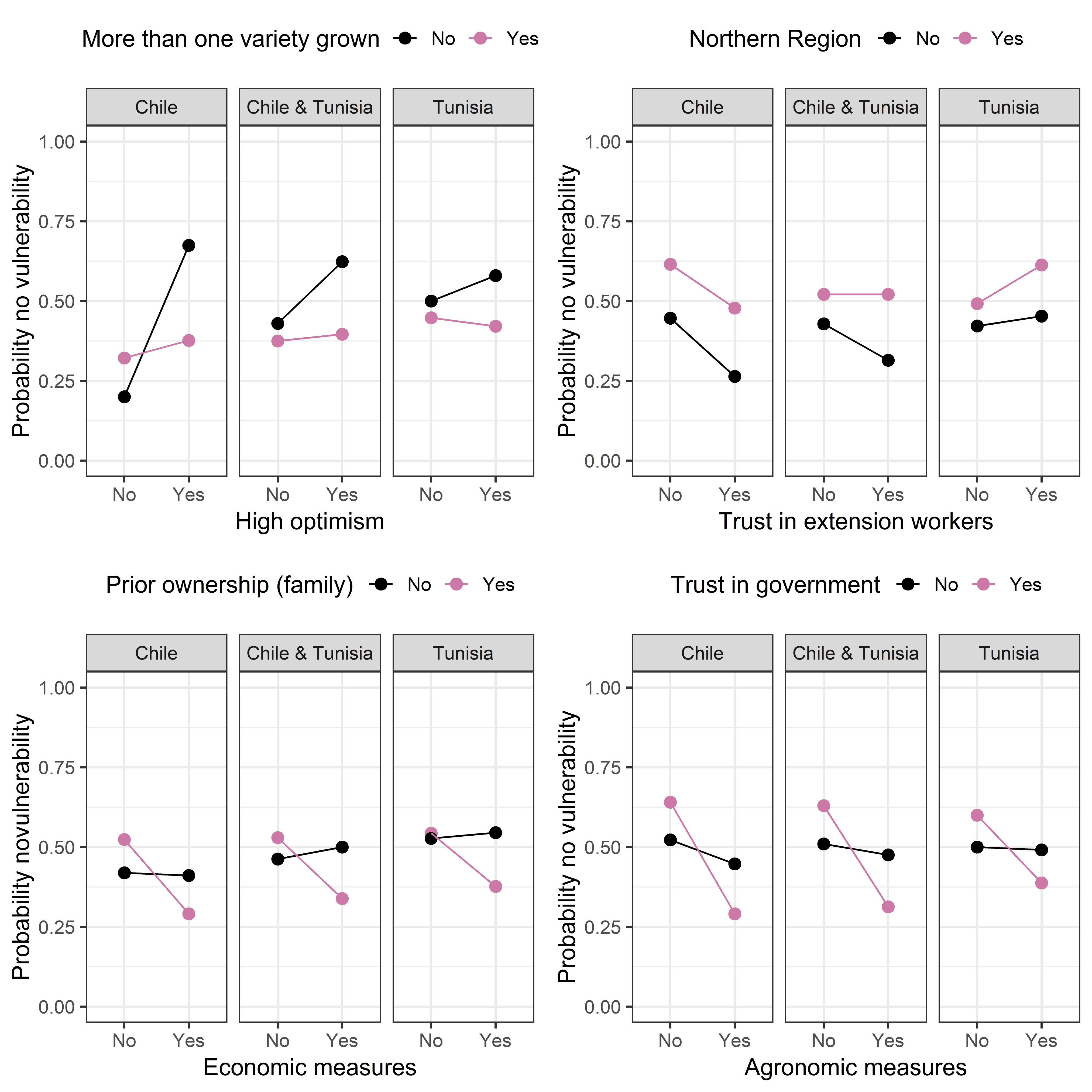}
\caption{Probability of not being vulnerable against increasing summer temperature based on the categories of the four most important interaction effects found in mb int and 2-boost. Probabilities are based on classical logistic regression only using one interaction term at a time. The results are once stratified by country (Chile, Tunisia) and once estimated on the whole data.}
\label{fig:S2.2c}
\end{figure}

\begin{figure}[ht]
\centering
\includegraphics[width=\linewidth]{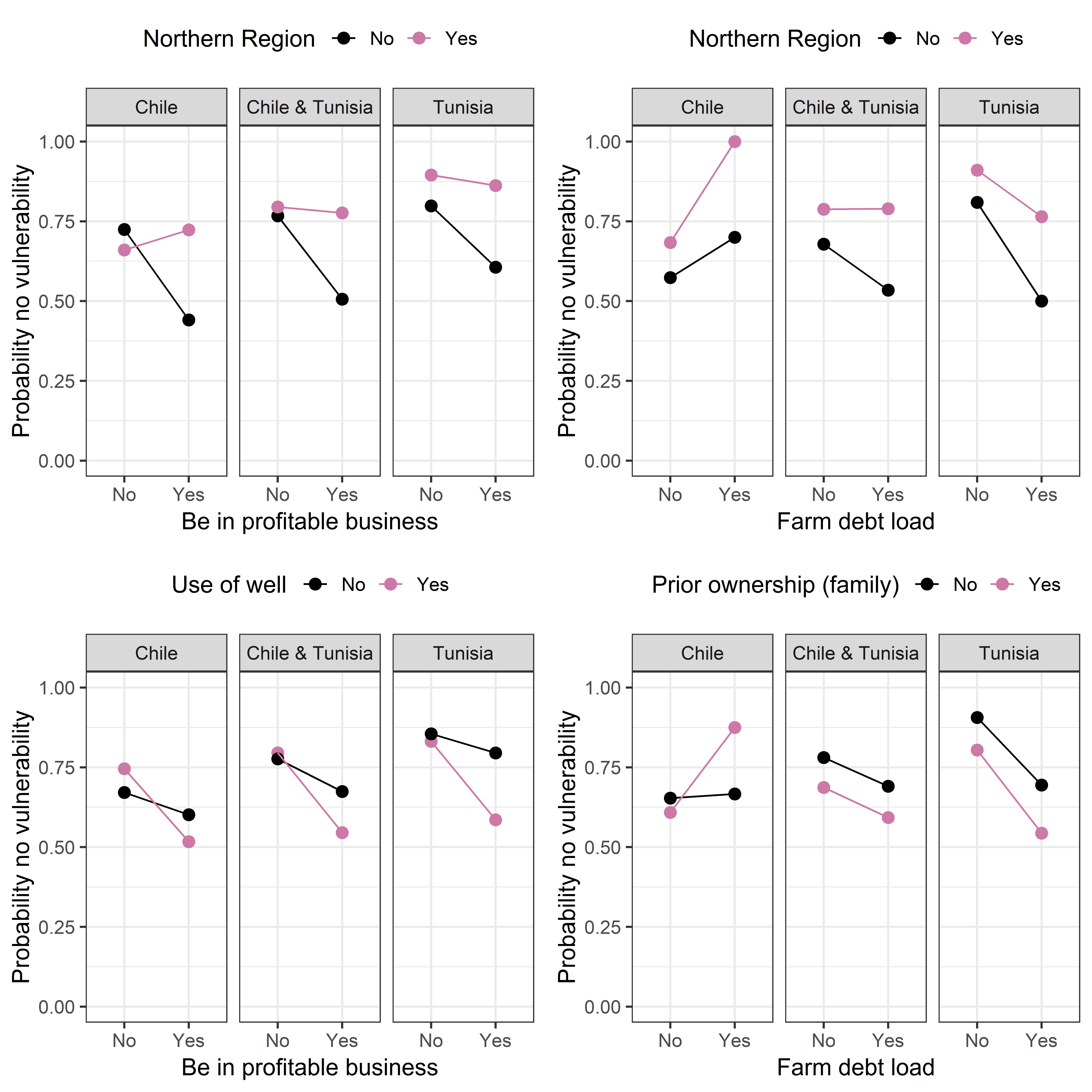}
\caption{Probability of not being vulnerable against increasing winter temperature based on the categories of the four most important interaction effects found in mb int and 2-boost. Probabilities are based on classical logistic regression only using one interaction term at a time. The results are once stratified by country (Chile, Tunisia) and once estimated on the whole data.}
\label{fig:S2.3c}
\end{figure}

\begin{figure}[ht]
\centering
\includegraphics[width=\linewidth]{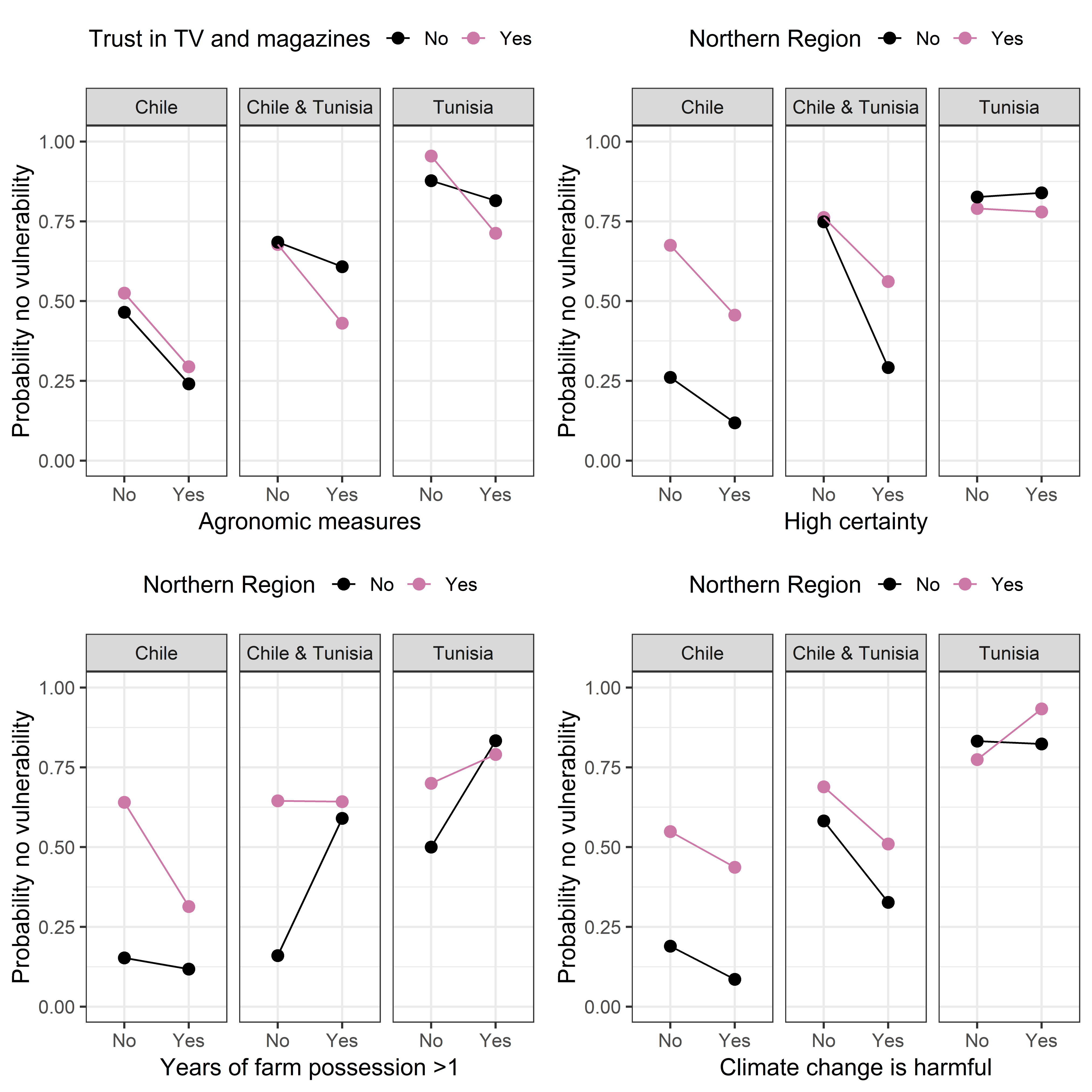}
\caption{Probability of not being vulnerable against decreasing rainfall based on the categories of the four most important interaction effects found in mb int and 2-boost. Probabilities are based on classical logistic regression only using one interaction term at a time. The results are once stratified by country (Chile, Tunisia) and once estimated on the whole data.}
\label{fig:S2.4c}
\end{figure}

\begin{figure}[ht]
\centering
\includegraphics[width=\linewidth]{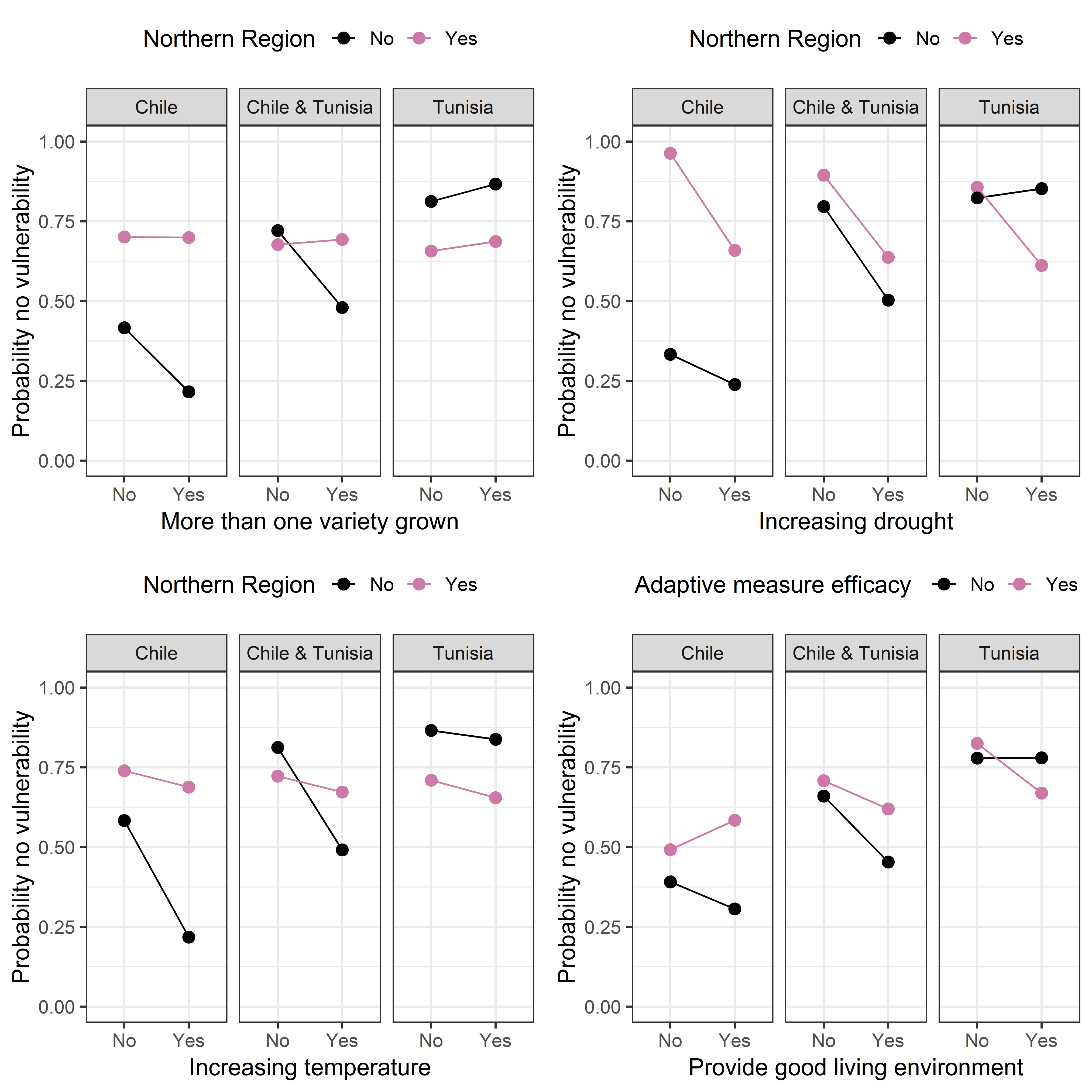}
\caption{Probability of not being vulnerable against drought based on the categories of the four most important interaction effects found in mb int and 2-boost. Probabilities are based on classical logistic regression only using one interaction term at a time. The results are once stratified by country (Chile, Tunisia) and once estimated on the whole data.}
\label{fig:S2.5c}
\end{figure}

\begin{figure}[ht]
\centering
\includegraphics[width=\linewidth]{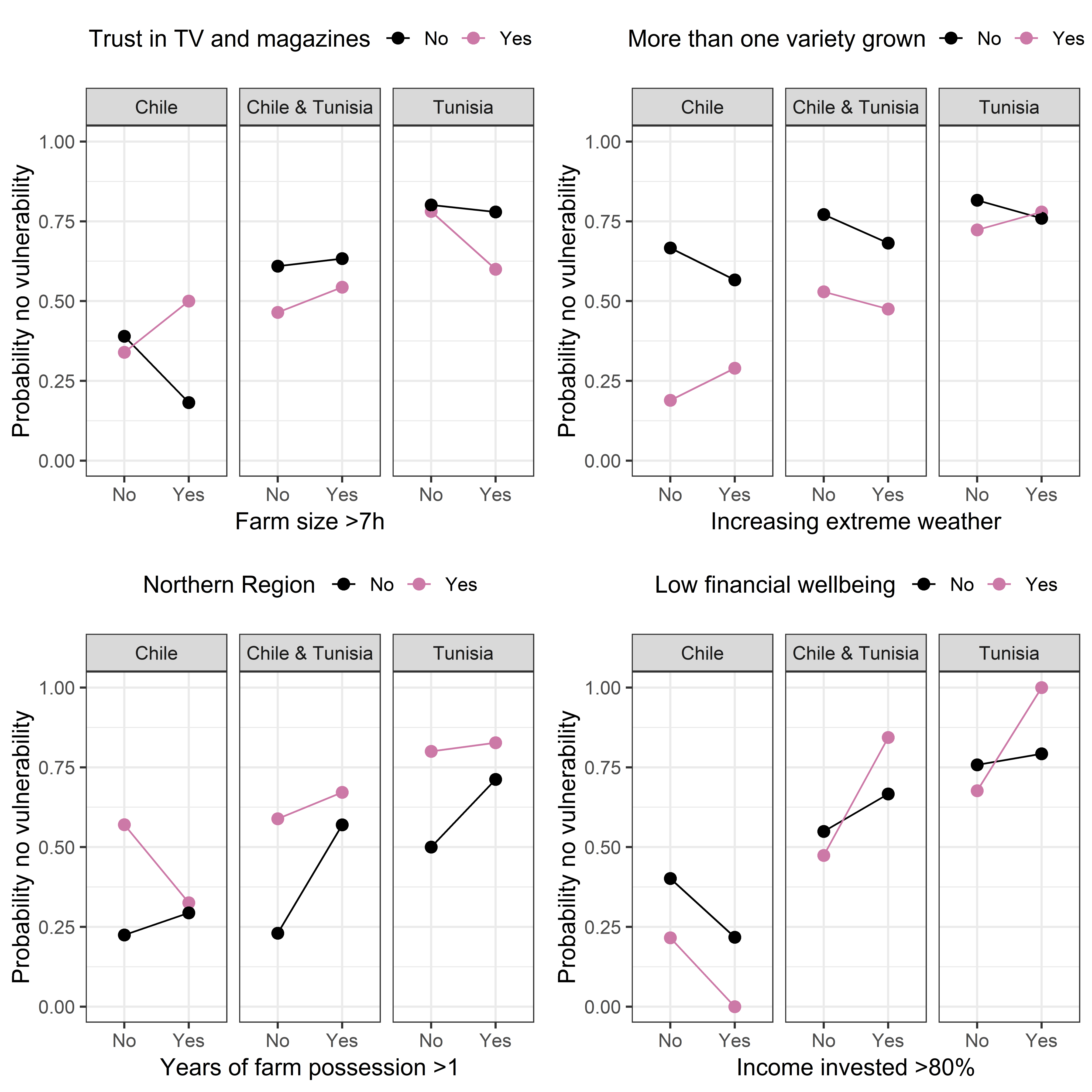}
\caption{Probability of not being vulnerable against extreme weather based on the categories of the four most important interaction effects found in mb int and 2-boost. Probabilities are based on classical logistic regression only using one interaction term at a time. The results are once stratified by country (Chile, Tunisia) and once estimated on the whole data. }
\label{fig:S2.6c}
\end{figure}

In the predictability section, we have already seen some differences between the two-step and the parallel estimation for interaction effects. For predictions, only models trained on the training data were used for model evaluation on the test data. For the variable importance in Figure \ref{fig:varimp_interact} and Sparsity in Table \ref{tab:sparsity_int} the whole data was used. The parallel estimation selected only interaction effects and no main effects (individual variables), whereas the two-step estimation selected both. \\  Referring to Table \ref{tab:sparsity_int} it becomes apparent that the selection of variables differs substantially. Overall, the two-step estimation in 2-boost yields much fewer interactions. For summer temperature vulnerability, no interaction term was selected, whereas for the parallel estimation, 13 interaction effects were selected. For decreasing rainfall vulnerability the differences are also substantial. The two-step estimation selected only one interaction, namely the one between Agronomic measures and trust in TV was selected and mb int selected 48. For drought vulnerability, the difference was the smallest with 27 interactions for the parallel and 16 for the two-step estimation. The percentage of selected interactions was four out of five times below one percent for 2-boost and for mb int it was above one percent four out of five times. \\
Not only does the sparsity differ, but also the selected interactions themselves. Referring to Figure \ref{fig:varimp_interact}, for winter temperature vulnerability the two interactions "Natural assets"-"Be profitable business" and "Country"-"Farm debt load" have high relative importance based on both models. But apart from those two, there is almost no overlap. For example for decreasing rainfall vulnerability, the only selected interaction between "Agronomic measures"-"Trust in TV" has a relative importance of 1 based on 2-boost and is not selected based on mb int, which in turn selected 48 other interactions. \\
In Figures \ref{fig:S2.2c}, \ref{fig:S2.3c}, \ref{fig:S2.4c}, \ref{fig:S2.5c}, \ref{fig:S2.6c} we plotted the four most important interaction effects for each of the vulnerabilities found in mb int and 2-boost based on a classical logistic regression only using one interaction term at a time. There, the probability of no vulnerability is plotted based on the joint categories of the interaction. This is done once for the data in Chile, Tunisia, and the whole data. Exemplary, we interpret the two common interaction effects "Natural assets"-"Be profitable business" and "Country"-"Farm debt load" for winter temperature vulnerability, which was selected by both models. In the northern region of Chile, having compared to not having the goal of being a profitable business is associated with a higher probability of not being vulnerable to increasing winter temperatures. In the southern Region of Chile, the association is reversed, meaning that having compared to not having the goal of being a profitable business is associated with a lower probability of not being vulnerable against increasing winter temperatures. In Tunisia, in both regions, the association of having the goal of being a profitable business is negative but more negative in the Southern region compared to the Northern region. Based on the interaction term "Country"-"Farm debt load", high farm debt load has a positive association with the probability of not being vulnerable to increasing winter temperature, where the association is negative in Tunisia. The positive association in Chile is stronger in the northern region and the negative association in Tunisia is stronger in the southern region.

\section*{Discussion}
The results indicate that the vulnerability of farmers in Chile and Tunisia against climate hazards can be predicted with the interpretable boosting algorithms and their variations by the variables and groups of variables used in the analysis. All models performed variable selection. The highest predictive power measured in AUC was achieved for vulnerability against decreasing rainfall and the lowest for summer temperature increases regardless of the type of boosting approach. For predicting summer temperature vulnerability the sparse group boosting outperformed all other models indicating that there may be underlying latent variables that cause the effects rather than the individual variables. The group variable importance mainly points to economic and biophysical assets including adaptive measures which may be an underlying determinant for summer temperature vulnerability. The variable importance strongly points to Natural assets consisting of the four different regions in Chile and Tunisia, which are a main determinant of all types of vulnerability. This indicates strong within and between country differences. The interaction analyses also confirm the importance of regionality, as some effects are strongly modulated by Country and North-South comparisons. The modulated effect of debt load by region may be an indication of economic differences between regions and closeness to bigger cities or could be a result of the different climatic zones.  \\
Even though there are strong interaction effects present in the data as seen in the univariate interaction analysis, it is not a simple task to transfer their presence into higher predictive power in a high-dimensional setting. This becomes apparent since the model including interactions base-learners additionally to the main effects performed worse than the same model without interactions in all cases. One of the reasons is probably overfitting, as the number of parameters to estimate exceeds the number of variables by a factor of over four. The result was that the interaction model did not include any main effects and only interactions. We believe that this issue of overfitting becomes more systematic in high-dimensional data than purely random because there if there are $p$ variables in the dataset, then there are $\mathcal{O}(p^2)$ possible interaction terms. So, with increasing $p$, the chance of selecting an interaction term over a main effect increases with regardless of the actual effect sizes. This implicit interaction selection bias could be addressed successfully by the proposed two-step boosting approach. The two-step boosting yielded higher predictive power and a higher degree of sparsity with fewer interactions being present in the resulting model. This leads us to believe that this approach is superior to the "classical" parallel estimation by including interaction terms in the main model formula in boosting. The only drawback we see is, that one has to estimate two models instead of just one which slightly increased the programming effort and reduces the potential for further parallelization as the models are fitted sequentially and not in parallel. However, it is common practice and in line with the principle of sparsity to always fit one model that contains only individual variables if one wants to do an interaction analysis \cite{aguinis_best-practice_2010}. In this case, the two-step boosting is also computationally more efficient because one can build upon the first model and avoid having to refit the main effect.     \\
In environmental research, consistently finding associations in high-dimensional datasets requires new methods to advance knowledge. These new methods allow more flexibility but often come at the cost of classical statistical inference, including p-values and estimations of standard errors as in the case of boosting \cite{hofner_gamboostlss_2014}. \\
Often, there are multiple plausible explanations for a phenomenon. The here proposed methods can enable direct comparison of a large number of explanations, estimating their explanatory importance for the outcome. This approach can accelerate understanding, particularly for newer phenomena like climate change, by gathering all variables that may be associated with the outcome and sampling observations for them. Starting with a relatively small sample size, one can estimate the relative importance of hypotheses and prioritize future research based on the results. \\
Using an apriori interpretable method, such as those previously described, provides the great advantage of being able to assess the predictability of a given set of explanations for an outcome. In contrast, post-hoc interpretability tools applied to a black box provide only a simplified explanation of how black-box predictions may be derived, without being able to assess how good the explanations themselves are at predicting the outcome. \\

\section*{Funding}
This research was conducted within the project “Phenological And Social Impacts of Temperature Increase – climatic consequences for fruit production in Tunisia, Chile and Germany” (PASIT; grant number 031B0467B of the German Federal Ministry of Education and Research). Open Access funding was enabled by LMU. Additional funding was provided by dtec.bw funded by NextGenerationEU.

\bibliography{references}

\begin{thebibliography}{10}
\urlstyle{rm}
\expandafter\ifx\csname url\endcsname\relax
  \def\url#1{\texttt{#1}}\fi
\expandafter\ifx\csname urlprefix\endcsname\relax\def\urlprefix{URL }\fi
\expandafter\ifx\csname doiprefix\endcsname\relax\def\doiprefix{DOI: }\fi
\providecommand{\bibinfo}[2]{#2}
\providecommand{\eprint}[2][]{\url{#2}}

\bibitem{li_data_2021}
\bibinfo{author}{Li, B.}, \bibinfo{author}{Chakraborty, S.},
  \bibinfo{author}{Weindorf, D.~C.} \& \bibinfo{author}{Yu, Q.}
\newblock \bibinfo{journal}{\bibinfo{title}{Data {Integration} {Using}
  {Model}-{Based} {Boosting}}}.
\newblock {\emph{\JournalTitle{SN Computer Science}}}
  \textbf{\bibinfo{volume}{2}}, \bibinfo{pages}{400},
  \doiprefix\url{10.1007/s42979-021-00797-0} (\bibinfo{year}{2021}).

\bibitem{obster_factors_2023}
\bibinfo{author}{Obster, F.}, \bibinfo{author}{Bohle, H.} \&
  \bibinfo{author}{Pechan, P.~M.}
\newblock \bibinfo{title}{Factors other than climate change are currently more
  important in predicting how well fruit farms are doing financially},
  \doiprefix\url{10.48550/arXiv.2301.07685} (\bibinfo{year}{2023}).
\newblock \bibinfo{note}{ArXiv:2301.07685 [cs, stat]}.

\bibitem{obster_improving_2023}
\bibinfo{author}{Obster, F.}, \bibinfo{author}{Brand, J.},
  \bibinfo{author}{Ciolacu, M.} \& \bibinfo{author}{Humpe, A.}
\newblock \bibinfo{journal}{\bibinfo{title}{Improving {Boosted} {Generalized}
  {Additive} {Models} with {Random} {Forests}: {A} {Zoo} {Visitor} {Case}
  {Study} for {Smart} {Tourism}}}.
\newblock {\emph{\JournalTitle{Procedia Computer Science}}}
  \textbf{\bibinfo{volume}{217}}, \bibinfo{pages}{187--197},
  \doiprefix\url{10.1016/j.procs.2022.12.214} (\bibinfo{year}{2023}).

\bibitem{freund_decision-theoretic_1997}
\bibinfo{author}{Freund, Y.} \& \bibinfo{author}{Schapire, R.~E.}
\newblock \bibinfo{journal}{\bibinfo{title}{A {Decision}-{Theoretic}
  {Generalization} of {On}-{Line} {Learning} and an {Application} to
  {Boosting}}}.
\newblock {\emph{\JournalTitle{Journal of Computer and System Sciences}}}
  \textbf{\bibinfo{volume}{55}}, \bibinfo{pages}{119--139},
  \doiprefix\url{10.1006/jcss.1997.1504} (\bibinfo{year}{1997}).

\bibitem{jennifer_feature_2022}
\bibinfo{author}{Jennifer, J.~J.}
\newblock \bibinfo{journal}{\bibinfo{title}{Feature elimination and comparison
  of machine learning algorithms in landslide susceptibility mapping}}.
\newblock {\emph{\JournalTitle{Environmental Earth Sciences}}}
  \textbf{\bibinfo{volume}{81}}, \bibinfo{pages}{489},
  \doiprefix\url{10.1007/s12665-022-10620-5} (\bibinfo{year}{2022}).

\bibitem{froeschke_spatio-temporal_2011}
\bibinfo{author}{Froeschke, J.~T.} \& \bibinfo{author}{Froeschke, B.~F.}
\newblock \bibinfo{journal}{\bibinfo{title}{Spatio-temporal predictive model
  based on environmental factors for juvenile spotted seatrout in {Texas}
  estuaries using boosted regression trees}}.
\newblock {\emph{\JournalTitle{Fisheries Research}}}
  \textbf{\bibinfo{volume}{111}}, \bibinfo{pages}{131--138},
  \doiprefix\url{10.1016/j.fishres.2011.07.008} (\bibinfo{year}{2011}).

\bibitem{tibshirani_regression_1996}
\bibinfo{author}{Tibshirani, R.}
\newblock \bibinfo{journal}{\bibinfo{title}{Regression {Shrinkage} and
  {Selection} via the {Lasso}}}.
\newblock {\emph{\JournalTitle{Journal of the Royal Statistical Society. Series
  B (Methodological)}}} \textbf{\bibinfo{volume}{58}},
  \bibinfo{pages}{267--288} (\bibinfo{year}{1996}).

\bibitem{zhao_boosted_2004}
\bibinfo{author}{Zhao, P.} \& \bibinfo{author}{Yu, B.}
\newblock \bibinfo{title}{Boosted {Lasso}}.
\newblock \bibinfo{type}{Tech. Rep.}, \bibinfo{institution}{CALIFORNIA UNIV
  BERKELEY DEPT OF STATISTICS} (\bibinfo{year}{2004}).
\newblock \bibinfo{note}{Section: Technical Reports}.

\bibitem{simon_sparse-group_2013}
\bibinfo{author}{Simon, N.}, \bibinfo{author}{Friedman, J.},
  \bibinfo{author}{Hastie, T.} \& \bibinfo{author}{Tibshirani, R.}
\newblock \bibinfo{journal}{\bibinfo{title}{A {Sparse}-{Group} {Lasso}}}.
\newblock {\emph{\JournalTitle{Journal of Computational and Graphical
  Statistics}}}  (\bibinfo{year}{2013}).

\bibitem{pechan_climate_2023}
\bibinfo{author}{Pechan, P.~M.}, \bibinfo{author}{Obster, F.},
  \bibinfo{author}{Marchioro, L.} \& \bibinfo{author}{Bohle, H.}
\newblock \bibinfo{journal}{\bibinfo{title}{Climate change impact on fruit farm
  operations in {Chile} and {Tunisia}.}}
\newblock {\emph{\JournalTitle{agriRxiv}}} \textbf{\bibinfo{volume}{2023}},
  \bibinfo{pages}{20230025166}, \doiprefix\url{10.31220/agriRxiv.2023.00171}
  (\bibinfo{year}{2023}).
\newblock \bibinfo{note}{Publisher: CABI International agriRxiv}.

\bibitem{pechan_reducing_2023}
\bibinfo{author}{Pechan, P.~M.}, \bibinfo{author}{Bohle, H.} \&
  \bibinfo{author}{Obster, F.}
\newblock \bibinfo{journal}{\bibinfo{title}{Reducing vulnerability of fruit
  orchards to climate change.}}
\newblock {\emph{\JournalTitle{agriRxiv}}} \textbf{\bibinfo{volume}{2023}},
  \bibinfo{pages}{20230025167}, \doiprefix\url{10.31220/agriRxiv.2023.00172}
  (\bibinfo{year}{2023}).
\newblock \bibinfo{note}{Publisher: CABI International agriRxiv}.

\bibitem{rstudio_team_rstudio_2020}
\bibinfo{author}{Team, R.}
\newblock \bibinfo{title}{{RStudio}: {Integrated} {Development} {Environment}
  for {R}} (\bibinfo{year}{2020}).

\bibitem{hofner_model-based_2014}
\bibinfo{author}{Hofner, B.}, \bibinfo{author}{Mayr, A.},
  \bibinfo{author}{Robinzonov, N.} \& \bibinfo{author}{Schmid, M.}
\newblock \bibinfo{journal}{\bibinfo{title}{Model-based boosting in {R}: a
  hands-on tutorial using the {R} package mboost}}.
\newblock {\emph{\JournalTitle{Computational Statistics}}}
  \textbf{\bibinfo{volume}{29}}, \bibinfo{pages}{3--35},
  \doiprefix\url{10.1007/s00180-012-0382-5} (\bibinfo{year}{2014}).

\bibitem{hothorn_mboost_2022}
\bibinfo{author}{Hothorn, T.}, \bibinfo{author}{Buehlmann, P.},
  \bibinfo{author}{Kneib, T.}, \bibinfo{author}{Schmid, M.} \&
  \bibinfo{author}{Hofner, B.}
\newblock \bibinfo{journal}{\bibinfo{title}{mboost: {Model}-{Based}
  {Boosting}}}.
\newblock {\emph{\JournalTitle{CRAN}}}  (\bibinfo{year}{2022}).

\bibitem{buhlmann_boosting_2007}
\bibinfo{author}{Bühlmann, P.} \& \bibinfo{author}{Hothorn, T.}
\newblock \bibinfo{journal}{\bibinfo{title}{Boosting {Algorithms}:
  {Regularization}, {Prediction} and {Model} {Fitting}}}.
\newblock {\emph{\JournalTitle{Statistical Science}}}
  \textbf{\bibinfo{volume}{22}}, \bibinfo{pages}{477--505},
  \doiprefix\url{10.1214/07-STS242} (\bibinfo{year}{2007}).

\bibitem{friedman_greedy_2001}
\bibinfo{author}{Friedman, J.~H.}
\newblock \bibinfo{journal}{\bibinfo{title}{Greedy function approximation: {A}
  gradient boosting machine.}}
\newblock {\emph{\JournalTitle{The Annals of Statistics}}}
  \textbf{\bibinfo{volume}{29}}, \bibinfo{pages}{1189--1232},
  \doiprefix\url{10.1214/aos/1013203451} (\bibinfo{year}{2001}).

\bibitem{tutz_boosting_2007}
\bibinfo{author}{Tutz, G.} \& \bibinfo{author}{Binder, H.}
\newblock \bibinfo{journal}{\bibinfo{title}{Boosting ridge regression}}.
\newblock {\emph{\JournalTitle{Computational Statistics \& Data Analysis}}}
  \textbf{\bibinfo{volume}{51}}, \bibinfo{pages}{6044--6059},
  \doiprefix\url{10.1016/j.csda.2006.11.041} (\bibinfo{year}{2007}).

\bibitem{agarwal_verifying_2011}
\bibinfo{author}{Agarwal, N.~K.}
\newblock \bibinfo{journal}{\bibinfo{title}{Verifying survey items for
  construct validity: {A} two-stage sorting procedure for questionnaire design
  in information behavior research}}.
\newblock {\emph{\JournalTitle{Proceedings of the American Society for
  Information Science and Technology}}} \textbf{\bibinfo{volume}{48}},
  \bibinfo{pages}{1--8}, \doiprefix\url{10.1002/meet.2011.14504801166}
  (\bibinfo{year}{2011}).

\bibitem{obster_sparse-group_2022}
\bibinfo{author}{Obster, F.} \& \bibinfo{author}{Heumann, C.}
\newblock \bibinfo{title}{Sparse-group boosting -- {Unbiased} group and
  variable selection}, \doiprefix\url{10.48550/arXiv.2206.06344}
  (\bibinfo{year}{2022}).
\newblock \bibinfo{note}{ArXiv:2206.06344 [stat]}.

\bibitem{pybus_hierarchical_2015}
\bibinfo{author}{Pybus, M.} \emph{et~al.}
\newblock \bibinfo{journal}{\bibinfo{title}{Hierarchical boosting: a
  machine-learning framework to detect and classify hard selective sweeps in
  human populations}}.
\newblock {\emph{\JournalTitle{Bioinformatics}}} \textbf{\bibinfo{volume}{31}},
  \bibinfo{pages}{3946--3952}, \doiprefix\url{10.1093/bioinformatics/btv493}
  (\bibinfo{year}{2015}).

\bibitem{wang_hierarchical_2016}
\bibinfo{author}{Wang, C.}, \bibinfo{author}{Wu, Y.} \& \bibinfo{author}{Liu,
  Z.}
\newblock \bibinfo{title}{Hierarchical boosting for transfer learning with
  multi-source}.
\newblock In \emph{\bibinfo{booktitle}{Proceedings of the {International}
  {Conference} on {Artificial} {Intelligence} and {Robotics} and the
  {International} {Conference} on {Automation}, {Control} and {Robotics}
  {Engineering}}}, {ICAIR}-{CACRE} '16, \bibinfo{pages}{1--5},
  \doiprefix\url{10.1145/2952744.2952756} (\bibinfo{publisher}{Association for
  Computing Machinery}, \bibinfo{address}{New York, NY, USA},
  \bibinfo{year}{2016}).

\bibitem{yang_feature_2019}
\bibinfo{author}{Yang, F.} \emph{et~al.}
\newblock \bibinfo{title}{Feature {Pyramid} and {Hierarchical} {Boosting}
  {Network} for {Pavement} {Crack} {Detection}},
  \doiprefix\url{10.48550/arXiv.1901.06340} (\bibinfo{year}{2019}).
\newblock \bibinfo{note}{ArXiv:1901.06340 [cs]}.

\bibitem{valentini_hierarchical_2014}
\bibinfo{author}{Valentini, G.}
\newblock \bibinfo{journal}{\bibinfo{title}{Hierarchical {Ensemble} {Methods}
  for {Protein} {Function} {Prediction}}}.
\newblock {\emph{\JournalTitle{ISRN Bioinformatics}}}
  \textbf{\bibinfo{volume}{2014}}, \bibinfo{pages}{901419},
  \doiprefix\url{10.1155/2014/901419} (\bibinfo{year}{2014}).

\bibitem{aguinis_best-practice_2010}
\bibinfo{author}{Aguinis, H.} \& \bibinfo{author}{Gottfredson, R.~K.}
\newblock \bibinfo{journal}{\bibinfo{title}{Best-practice recommendations for
  estimating interaction effects using moderated multiple regression}}.
\newblock {\emph{\JournalTitle{Journal of Organizational Behavior}}}
  \textbf{\bibinfo{volume}{31}}, \bibinfo{pages}{776--786},
  \doiprefix\url{10.1002/job.686} (\bibinfo{year}{2010}).

\bibitem{hofner_gamboostlss_2014}
\bibinfo{author}{Hofner, B.}, \bibinfo{author}{Mayr, A.} \&
  \bibinfo{author}{Schmid, M.}
\newblock \bibinfo{journal}{\bibinfo{title}{{gamboostLSS}: {An} {R} {Package}
  for {Model} {Building} and {Variable} {Selection} in the {GAMLSS}
  {Framework}}}.
\newblock {\emph{\JournalTitle{arXiv:1407.1774 [stat]}}}
  (\bibinfo{year}{2014}).
\newblock \bibinfo{note}{ArXiv: 1407.1774}.

\end{thebibliography}

\section*{Author contributions statement}

P.P. and H.B. accumulated the data, F.O. performed machine learning and statistical modeling, and F.O. analysed the results. All authors reviewed the manuscript. 

\section*{Additional information}

To include, in this order: \textbf{Accession codes} (where applicable); \textbf{Competing interests} (mandatory statement).

\begin{table}[ht]
\centering
\begin{tabular}{|l|l|l|l|l|l|l|l|l|l|}
\hline
AUC sgb & AUC mb & AUC 2-boost &AUC mb int & outcome vulnerability
\\
\hline
0.656 & 0.619 & 0.608 & 0.587 & Summer temperature \\
\hline
0.707 & 0.708 & 0.713 & 0.705 & Winter temperature \\
\hline
0.852 & 0.852 & 0.852 & 0.500 & Decreasing rainfall \\
\hline
0.768 & 0.768 & 0.768 & 0.500 & Drought\\
\hline
0.776 & 0.778 & 0.783 & 0.773 & Extreme weather \\
\hline
\end{tabular}
\caption{\label{tab:auc_no_vul} AUC values for the sparse group boosting (sgb), component-wise boosting (mb), parallel boosting with interaction (mb int) and two-step boosting with interactions (2-boost) for all vulnerability outcomes evaluated on the test data.
}
\end{table}

\begin{table}[ht]
\centering
\begin{tabular}{|l|l|l|l|l|l|l|l|l|l|}
\hline
model & Number selected interaction terms & 1-Sparsity in percent & outcome vulnerability
\\
\hline
mb int & 13 & 0.95 & Summer temperature \\
\hline
2-boost & 0 & 0 & Summer temperature \\
\hline
mb int & 38 & 2.78 & Winter temperature \\
\hline
2-boost & 12 & 0.88 & Winter temperature \\
\hline
mb int & 48 & 3.51 & Decreasing rainfall \\
\hline
2-boost & 1 & 0.07 & Decreasing rainfall \\
\hline
mb int & 27 & 1.98 & Drought\\
\hline
2-boost & 16 & 1.17 & Drought\\
\hline
mb int & 32 & 2.34 & Extreme weather \\
\hline
2-boost & 10 & 0.73 & Extreme weather \\
\hline
\end{tabular}
\caption{\label{tab:sparsity_int} Comparison of the number of selected interaction terms based on two-step estimation (2-boost) and the parallel estimation (mb int) and the percentage of selected interactions (1-Sparsity) of the 1366 interaction terms.}
\end{table}

\begin{table}[ht]
\centering
\begin{tabular}{|l|l|l|l|l|l|l|l|l|l|}
\hline
variable&category & n \\
\hline
No summer temperature vulnerability & yes & 358 \\
\hline
No winter temperature vulnerability & yes & 579\\
\hline
No decreasing rainfall vulnerability & yes & 451 \\
\hline
No drought vulnerability & yes & 492\\
\hline
No Extreme weather vulnerability & yes & 453 \\
\hline
\end{tabular}
\caption{\label{tab:outcomes} Overview over outcome variables. Financial vulnerability against climate hazards. The "n" column gives the number of farmers who are not financially vulnerable to each of the hazards.}
\end{table}

\begin{table}[ht]
\centering
\begin{tabular}{|l|l|l|l|l|l|l|l|l|l|}
\hline
variable name & category & n & group name
\\
\hline
\textbf{Agronomic measures} & yes & 647 & Biophysical asset group \\
\hline
\textbf{Economic measures} & yes & 464 & Biophysical asset group\\
\hline
Use of river irrigation & yes & 138 & Biophysical asset group\\
\hline
\textbf{Use of well irrigation} & yes & 231 & Biophysical asset group\\
\hline
Farm size & yes & 283 & Biophysical asset group\\
\hline
Orchard size & yes & 318 & Biophysical asset group\\
\hline
More than one variety grown & yes & 508 & Biophysical asset group\\
\hline
Other products & yes & 571 & Biophysical asset group\\
\hline
\textbf{Technological measures} & yes & 721 & Biophysical asset group\\
\hline
\textbf{Increasing temperature} & yes & 629 & Climate experience group\\
\hline
\textbf{Decreasing rainfall} & yes & 659 & Climate experience group\\
\hline
\textbf{Increasing drought} & yes & 671 & Climate experience group\\
\hline
\textbf{Increasing extreme weather} & yes & 542 & Climate experience group\\
\hline
\textbf{Income invested >80 Percent} & yes & 137 & Economic asset group\\
\hline
Income invested <40 percent & yes & 358 & Economic asset group\\
\hline
High financial wellbeing & yes & 346 & Economic asset group\\
\hline
Low financial wellbeing & yes & 148 & Economic asset group\\
\hline
\textbf{Farm debt load} & high & 96 & Economic asset group\\
\hline
Dependent on farm & yes & 528 & Economic asset group\\
\hline
Family farm engagement & yes & 203 & Economic asset group\\
\hline
\textbf{Adaptive measures efficacy} & high & 490 & Efficacy group\\
\hline
Work independent & yes & 635 & Goals group\\
\hline
\textbf{Keep tradition alive} & yes & 460 & Goals group\\
\hline
Provide good living environment & yes & 466 & Goals group\\
\hline
Be in profitable business & yes & 320 & Goals group\\
\hline
Climate change is harmful & yes & 258 & Harm group\\
\hline
\textbf{High optimism} & yes & 446 & Harm group\\
\hline
High certainty & yes & 470 & Harm group\\
\hline
Climate threatens farm & yes & 629 & Harm group\\
\hline
Climate risks > benefits & yes & 648 & Harm group\\
\hline
Climate change acceptance & yes & 676 & Human asset group\\
\hline
Human cause climate change & yes & 685 & Human asset group\\
\hline
Climate extremes & yes & 755 & Human asset group\\
\hline
\textbf{Age > 50} & yes & 438 & Human asset group\\
\hline
\textbf{Gender} & F & 121 & Human asset group\\
\hline
\textbf{Gender} & M & 680 & Human asset group\\
\hline
\textbf{Education} & yes & 459 & Human asset group\\
\hline
Years of farm possession & yes & 577 & Human asset group\\
\hline
Prior ownership (family) & yes & 399 & Human asset group\\
\hline
Years of farm managing & yes & 437 & Human asset group\\
\hline
\textbf{Natural assets} & CentralChile & 200 & Natural asset group\\
\hline
\textbf{Natural assets} & CentralTunisia & 200 & Natural asset group\\
\hline
\textbf{Natural assets} & NorthernTunisia & 201 & Natural asset group\\
\hline
\textbf{Natural assets} & SouthernChile & 200 & Natural asset group\\
\hline
Adapive measures near farms & 1 & 424 & Norms group\\
\hline
Adapive measures near farms & 2 & 151 & Norms group\\
\hline
Adapive measures near farms & 3 & 226 & Norms group\\
\hline
High optimism & yes & 446 & Perception group\\
\hline
\end{tabular}
\caption{\label{tab:variables} Overview over variables and groups. The 22 variables used as interaction variables (potential moderators) are bold. The number of observations within each category of each variable is in the n column. For binary variables, only one category is presented and the remaining category is "no" if the shown category is "yes" and "low" if the shown category is "high"}
\end{table}

\begin{table}[ht]
\centering
\begin{tabular}{|l|l|l|l|l|l|l|l|l|l|}
\hline
variable & category & n & group
\\
\hline
Use of newspapers & yes & 95 & Social asset group\\
\hline
Use of farming journals & yes & 161 & Social asset group\\
\hline
\textbf{Use of TV} & yes & 415 & Social asset group\\
\hline
Use of radio & yes & 219 & Social asset group\\
\hline
Use of internet & yes & 319 & Social asset group\\
\hline
Use of extension workers & yes & 346 & Social asset group\\
\hline
Use of government workers & yes & 166 & Social asset group\\
\hline
Use of neighbours & yes & 313 & Social asset group\\
\hline
Use of industry & yes & 192 & Social asset group\\
\hline
\textbf{Use of farm associations} & yes & 97 & Social asset group\\
\hline
Trust in newspapers & yes & 174 & Social asset group\\
\hline
Trust in farming journals & yes & 291 & Social asset group\\
\hline
\textbf{Trust in TV} & yes & 329 & Social asset group\\
\hline
Trust in radio & yes & 241 & Social asset group\\
\hline
Trust in internet & yes & 319 & Social asset group\\
\hline
Trust in extension workers & yes & 433 & Social asset group\\
\hline
Trust in government workers & yes & 268 & Social asset group\\
\hline
Trust in neighbours & yes & 319 & Social asset group\\
\hline
\textbf{Trust in industry} & yes & 215 & Social asset group\\
\hline
Trust in farm associations & yes & 184 & Social asset group\\
\hline
Trust in government institutions & yes & 312 & Social asset group\\
\hline
Trust in other farmers & yes & 351 & Social asset group\\
\hline
Trust in religion & yes & 317 & Social asset group\\
\hline
Trust in fate & yes & 360 & Social asset group\\
\hline
Crop damage near farms & yes & 643 & Spatial group\\
\hline
Crop damage farms in Country & yes & 673 & Spatial group\\
\hline
Climate change occurs & yes & 592 & Spatial group\\
\hline
\end{tabular}
\caption{\label{tab:variables_2} Overview over variables and groups continued. The 22 variables used as interaction variables (potential moderators) are bold. The number of observations within each category of each variable is in the n column. For binary variables, only one category is presented and the remaining category is "no" if the shown category is "yes" and "low" if the shown category is "high"}
\end{table}
\end{document}